# Automatic Normalization of Word Variations in Code-Mixed Social Media Text

Rajat Singh*, Nurendra Choudhary* and Manish Shrivastava

Language Technologies Research Centre (LTRC)
Kohli Center on Intelligent Systems (KCIS)
International Institute of Information Technology, Hyderabad, India
{rajat.singh,nurendra.choudhary}@research.iiit.ac.in
m.shrivastava@iiit.ac.in

**Abstract** Social media platforms such as Twitter and Facebook are becoming popular in multilingual societies. This trend induces portmanteau of South Asian languages with English. The blend of multiple languages as code-mixed data has recently become popular in research communities for various NLP tasks. Code-mixed data consist of anomalies such as grammatical errors and spelling variations. In this paper, we leverage the contextual property of words where the different spelling variation of words share similar context in a large noisy social media text. We capture different variations of words belonging to same context in an unsupervised manner using distributed representations of words. Our experiments reveal that preprocessing of the code-mixed dataset based on our approach improves the performance in state-of-the-art part-of-speech tagging (POS-tagging) and sentiment analysis tasks.

## 1 Introduction

Code-mixing is the embedding of linguistic units such as phrases, words or morphemes of one language into the utterance of another language, whereas code-switching refers to the co-occurrence of speech extracts belonging to two different grammatical systems. Prevalent use of social media platforms by multilingual speakers leads to an increase in the phenomenon of code-mixing and code-switching [5,10,3,6]. Here we refer both the scenarios as code-mixing. Hindi-English bilingual speakers generate an immense amount of code-mixed social media text (CSMT). [19] noted the complexity in analyzing CSMT stems from non-adherence to a formal grammar, spelling variations, lack of annotated data, inherent conversational nature of the text and code-mixing. Traditional tools presume texts to have a strict adherence to formal structure. Hence, unique natural language processing (NLP) tools for CSMT are required and should be improved upon.

Internet usage is steadily increasing in multilingual societies such as India, where there are 22 official languages at center and state level, out of which

---

* These authors have contributed equally to this work.

Hindi and English are most prevalent[1]. These multilingual populations actively use code-mixed language on social media to share their opinion. With over 400 million Indian population on Internet, which is predicted to double in next 5 years[2], we notice a huge potential for research in CSMT data.

On analyzing CSMT data, we find following ways in which CSMT data deviates from a formal standard form of the respective language:

– Informal transliteration: These variations are due to lack of a transliteration standard. Multilingual speakers tend to transliterate the lexicons directly from native script to roman, which lacks a formal transliteration method. Hence, it leads to many phonetic variations. For example, बहुत can be transliterated to *bahoot*, *bahout* or *bahut*, which may be based on socio-cultural factors like accent, dialect and region.
– Informal speech: These variations are not language specific. Speakers tend to write non standard spellings on social media. Usage of a non-formal speech leads to variations in spellings. For example, coooooool, gud, mistke or lappy.

Section 3 discusses the variations in more detail.

Unless a system captures these variations in code-mixed data, its performance will not be at par with corresponding systems on formal standard texts. In this paper, we present a novel approach of automating the normalization process of code-mixed informal text. We also compare the performances of current state-of-the-art CSMT sentiment analysis [12] and POS-tagging [4] tasks on CSMT data.

Section 2 discusses some relevant recent work and section 3 describes and explains these variations. Section 4 provides information about the datasets we use in this paper. Section 5 gives detailed methodology of our approach. The experiments and evaluations are presented in Section 6 and the conclusion is in Section 7.

## 2 Related Work

Analysis of code-mixed languages has recently gained interest owing to the rise of non-native English speakers. [15] normalized the code-mixed text by segregating Hindi and English words. Hindi-English (Hi-En) code-mixing allows ease-of-communication among speakers by providing a much wider variety of phrases and expressions. A common form of code-mixing is romanization [15], which refers to the transliteration from a different writing system to the roman script. But this freedom makes formal rules irrelevant, leading to more complexities in the NLP tasks pertaining to CSMT data, highlighted by [2,19,1].

Initiatives have been taken by shared tasks [14,17] for analysis of CSMT data. [18] used distributional representation to normalize English social media text

---

[1] http://www.thehindu.com/data/sanskrit-and-english-theres-no-competition/article6630269.ece
[2] https://yourstory.com/2017/06/india-internet-users-report/

Automatic Normalization of Word Variations 3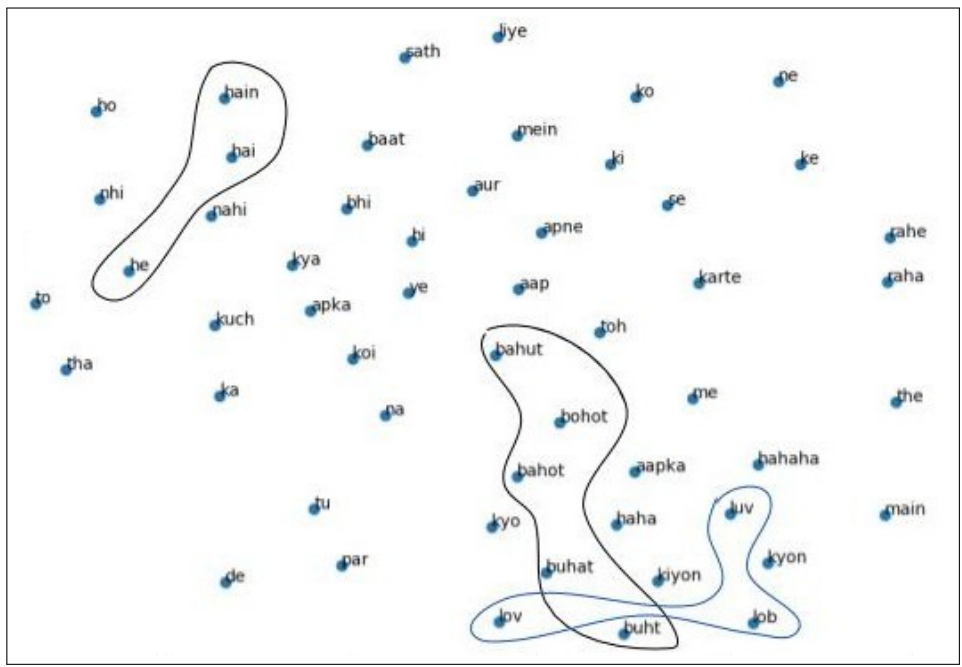

Figure 1: Most frequent 50 words with some clusters from CSMT data

by substituting spelling mistakes with their corresponding normal form. Deep learning based solutions [21,16] are also demonstrated for various NLP tasks. [15] provides a methodology for normalization of these variably spelled romanized words in a rule-based manner. However, giving an automated unsupervised machine learning model enables the system to be utilized across languages. This is important in case of other Indian Languages, which exhibit similar code-mixing and romanized behavior as Hindi.

## 3 Types of Variations

Informal variations in lexical forms have not been explored properly. Hence we provide our own nomenclature to better address the normalization process and to provide a discourse for further discussion.

As explained in Section 1, we observe informal variations of words in CSMT data. The context we explain the variations here is in Hindi-English per se. However, the approach is not language-specific.

- **Informal transliterations**: Lack of a transliteration standard implies that decisions of marking vowels and other sounds rely entirely on the user. This happens in case of romanizing Hindi which is written phonetically.

- **Long Vowel transliteration**: Users are found to be indicating vowel length in long vowels in many fashions. For example, the word खाया: is transliterated to *khaaya*, repeating the vowel twice or as *khAya*, using an upper case for the vowel character or just *khaya*, not indicating the length at all. Similar variations observed in साल: *saal*, *sAl*, *sal*,; मेरा: *meraa*, *merA*, *mera* ; आजा: *aajaa*, *AjA*, *aja*, *aaja* ; or in आपका: *aapka*, *apka*, *Apka*.
- **Borrowed words' pronunciations**: In case of Indian Languages, especially in Hindi, influence of Persian and English is observed. So when users transliterate foreign words, they rely on the pronunciation when transliterating. For example, some Persian sounds lack an exact counterpart in Hindi, because of which, both इज्जत *ijjat*, and इज़्ज़त *izzat* are observed. Same is observed in आज़ाद *azad* and आजाद *ajad*; ज़िंदाबाद *zindabad* and जिन्दाबाद *jindabad*. This is also observed in English to Hindi transliteration. For example, hospital may be written as (अस्पताल) *aspataal* or (होस्पिटल) *hospital*.
- **Accent and Dialect Phonetics**: Not all alternate pronunciations arise from different languages, they may be drawn from difference in dialects or accents between users and its transliteration reflects the user's utterance of the word. For example, श्री could be transliterated as both *Shree*, *Sree*, similarly शपत into *shapat* or *sapat*. Certain accents of Hindi speakers lack aspiration. For example, when transliterating खाया to *khaya* or *kaya*.
- **Double Consonants**: Similar to long vowels, Hindi contains words with double consonants where in speech, stress is given on respective sound. However it is observed that users use variants with or without repeating the respective consonant. For example, इज्जत: *izzat*, *izat* with stress on *z* or in स्वच्छता: *swachhta*, *swachta* with stress on *ch*.
- **Non phonetic writing**: Users transliterate differently despite having common pronunciation. The reason is that in English alphabets, multiple characters may have same pronunciation contextually. At the same time, same character may have different pronunciations. For example, k or q may have same pronunciations(king, queue) and hence वक्त could be transliterated as either *waqt* or *wakt*. Similarly in *ee* and *i* pronunciation like in case of *spree* and *ski* is observed. Hence, transliteration of श्री as *Shree* or *Shri* is noticed.

– **Informal speech**: These are variations in spellings which are caused by non formal speech. Following are some variations caused by the informal setting of social media.
  - **Elongation**: Although, speakers have knowledge of word spellings, some spellings are stretched to indicate certain sentiments, like of joy or excitement, word forms like coooooool, soooooo goood or noooo convey an emphasis which their respective correct spellings can not.
  - **SMS Language**: An interesting phenomena that is observed in informal social media is of conveying messages in limited number of characters, this style originates from early SMS applications, where messages had a size limit. A noticeable pattern in phrases like *gud nite*, *plz dnt do ths*

Automatic Normalization of Word Variations 5| Language Tags | Utterances (Training) | Utterances(Training) |
|---|---|---|
| Hindi-English | | |
| English | 6178 | 8553 |
| Hindi | 5546 | 411 |
| Others | 4231 | 2248 |
| Total | 15955 | 11212 |
| Bengali-English | | |
| English | 9973 | 5459 |
| Bengali | 8330 | 4671 |
| Others | 6335 | 3431 |
| Total | 24638 | 13561 |
| Tamil-English | | |
| English | 1969 | 819 |
| Tamil | 1716 | 1155 |
| Others | 630 | 281 |
| Total | 4315 | 2255 |

Table 1: Summary of Dataset(Utterances) used for POS-tagging

or *nerndr kb arhn hn?*(in Hindi) is that of dropping most vowels and retaining consonants just enough to guess the word.

- **Abbreviations**: In informal speech, speakers often use abbreviated forms of words, which translates to informal writing. For example, *lappy* for laptop.

| Language | Sentences (Training) | Sentences (Text) |
|---|---|---|
| Bengali-English | 2837 | 1459 |
| Hindi-English | 729 | 377 |
| Tamil-English | 639 | 279 |

Table 2: Summary of Dataset(Sentences) used for POS-tagging

| Data | Vocab | #Sentences |
|---|---|---|
| Twitter Corpus | 198025 | 500000 |
| Hi-En code-mixed data [12] | 7549 | 3879 |

Table 3: Statistics of the Hindi-English code-mixed data used to learn distributed representations and Hindi-English code-mixed dataset used for sentiment analysis

## 4 Dataset

We utilize Twitter as the source of code-mixed text by scrapping 500,000 tweets. We first collect 500 most frequent words in the romanized form of the target language pair (Hindi in our case of Hi-En code-mixed text). Multiple websites showcase most frequent words, including their roman transliterated forms in a particular language. We adopt wiktionary.com[3] for this task. Using these collected most frequent words, we create queries to fetch code-mixed tweets. These queries fetch 500,000 Hindi-English (Hi-En) code-mixed tweets using Twitter APIs[4]. To train a skip-gram model[9] with meaningful vector distribution, we need large data. We remove the punctuations and lowercase the text. These scraped tweets are the CSMT training data for the skip gram model in our preprocessing task. The statistics of this corpus is given in table 3. This technique easily extends to any code-mixed language pair.

| $\epsilon$ | error % | Average #Words per cluster |
|---|---|---|
| **0** | 1.2 | 3.7 |
| **1** | 8.3 | 4.5 |
| **2** | 28.1 | 7.2 |
| **3** | 47.7 | 11.4 |

Table 4: Error analysis of clusters using 500 random sample of words

| **Method** | Without our approach | | Proposed preprocessing | |
|---|---|---|---|---|
| | **A** | **F1** | **A** | **F1** |
| NBSVM(Unigram)[20] | 59.15% | 0.5335 | **60.38%** | **0.5421** |
| NBSVM(Uni+Bigram)[20] | 62.5% | 0.5375 | **63.43%** | **0.5566** |
| MNB(Unigram)[20] | 66.75% | 0.6143 | **67.23%** | **0.6325** |
| MNB(Uni+Bigram)[20] | 66.36% | 0.6046 | **68.85%** | **0.6367** |
| MNB(Tf-Idf)[20] | 63.53% | 0.4783 | **65.76%** | **0.5025** |
| SVM(Unigram)[11] | 57.6% | 0.5232 | **58.23%** | **0.5325** |
| SVM(Uni+Bigram)[11] | 52.96% | 0.3773 | **55.63%** | **0.4238** |
| Lexicon Lookup[15] | 51.15% | 0.252 | **53.28%** | **0.2745** |
| Char-LSTM[12] | 59.8% | 0.511 | **60.82%** | **0.5285** |
| Subword-LSTM[12] | 69.7% | 0.658 | **71.38%** | **0.6621** |

Table 5: Comparison of proposed preprocessing in Sentiment Analysis Task. A, F1 are Accuracy and F1-scores of the models respectively.

For sentiment analysis task, we consider the dataset (statistics in table 3) from [12] which is 4981 annotated sentences with 15% negative, 50% neutral and 35% positive comments. For POS-tagging of transliterated social media task we adopt dataset from the shared task conducted by Twelfth International Conference on Natural Language Processing (ICON-2015)[5]. Organizers released the code-mixed train and test set for English-Hindi, English-Bengali and English-Tamil language pairs. In table 1, we provide a summary of the dataset in terms of the utterances. The number of utterances have been recorded for both the train-ing and test data. In table 2, we present a statistics of the number of sentences for each pair of languages in training as well as test data.

---

[3] https://en.wiktionary.org/wiki/Wiktionary:Frequency_lists

[4] https://developer.twitter.com/en/docs/tweets/search/overview

[5] http://ltrc.iiit.ac.in/icon2015/contests.php

| | Without our approach | | Proposed Preprocessing | |
|---|---|---|---|---|
| **Language Pair** | **Baseline (Stanford Model)** | **CRF Model** | **Baseline (Stanford Model)** | **CRF Model** |
| Bengali-English | 60.05% | 75.22% | **62.27%** | **76.14%** |
| Hindi-English | 50.87% | 73.2% | **53.45%** | **75.24%** |
| Tamil-English | 61.02% | 64.83% | **63.38%** | **65.97%** |

Table 6: Comparison of proposed preprocessing on POS-tagging

## 5 Methodology

In this section, we discuss our methodology (shown in figure 2) for preprocessing the CSMT data for NLP tasks pertaining to CSMT.

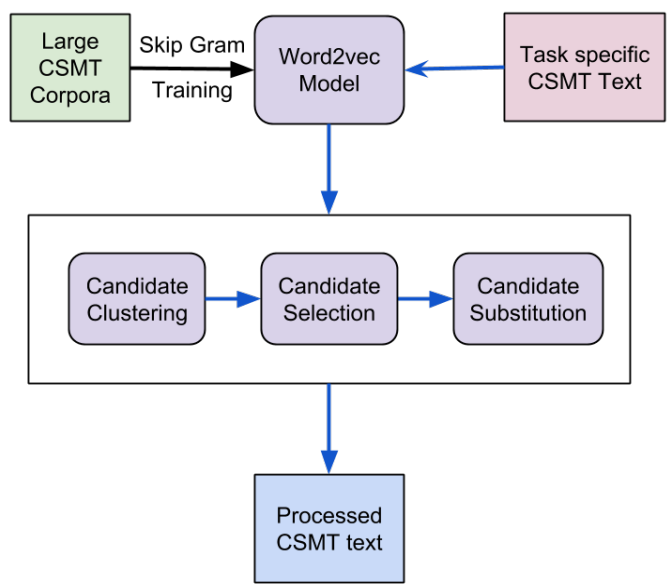

Figure 2: Methodology

| Standard Form | Meaning | Captured Variations | | | |
|---|---|---|---|---|---|
| खूबसूरत | beautiful | **khoobsurat** | khubsurat | khubsoorat | khbsurt |
| क्यूंकि | because | **kyunki** | kiyunki | kiyuki | kyuki |
| मेहरबानी | clemency | **meherbani** | meharbaani | meharbani | meherbanee |
| आपका | yours | **aapka** | apkaa | apka | apkA |

Table 7: Some example clusters with their respective parent candidate shown in bold.

### 5.1 Skip Gram distributional semantic modeling

We use the collected 500,000 Hi-En CSMT tweets to train Word2Vec embeddings, computed on a 300-dimensional embedding with a skip-gram-10 model[9]. We used the python gensim module [13] to train the representation. We utilized hierarchical sampling for the reduction in vocabulary count during training and used a minimum count of 5 occurrences for each word. This model will be used to extract candidate words and their related variations using clustering. The architecture discussed in this paper can also work with word vectors obtained using other techniques such as latent semantic indexing, convolutional neural networks, recurrent neural networks, etc.

### 5.2 Candidate Clustering

Skip-gram vectors give the representation of a word in the semantic space based on their context. Due to contextual similarity, variations of the same word will have similar vector representation. Also, it is observed, that the consonants of these variations are similar. Hence, we cluster the words based on a distance metric that captures both these properties. The similarity metric is formally defined below:

$$f(v1, v2) = \begin{cases} sim(vec(v1), vec(v2)) & \text{if EditDistance(v1,v2)} \leq \epsilon \\ 0 & \text{else} \end{cases} \quad (1)$$

where $v1$ and $v2$ are the two variations, *sim* is a vector similarity function (cosine similarity in our case), $vec(v)$ is a function that returns the skip-gram vector of $v$, $f(v1, v2)$ represents the overall similarity between $v1$ and $v2$, and the threshold of Levenshtein distance[7] (also known as Edit Distance). $\epsilon$ is set according to our requirement of the task. If $\epsilon$ is 0 then we will get small clusters with more accuracy but as we increase it further the accuracy will reduce but the coverage will increase with large clusters. We randomly sampled 500 words with their clusters and manually verified the variations of words and calculated the error percentage shown in table 4. We chose $\epsilon$ as 1 for better accuracy and increased coverage. While calculating Levenshtein distance for each pair in clusters we remove the vowels, repetition reduced to single character and digits substituted by their corresponding letters such as *2→to* and *4→for*. The $\epsilon$ varied from 0-2. This metric gives us the closest variations for the given word. They together form a cluster. Example clusters have been presented in table 7. Figure 1 showcase most frequent 50 words from the Hindi-English code-mixed large corpora plotted using Matplotlib library[6] and t-SNE[7] from the skip-gram model trained by large code-mixed corpora of Hindi-English language pair. t-SNE[8] helps in the dimensionality reduction to visualize 300 dimensions words.

### 5.3 Substitution

Finally, in the given CSMT dataset for the task, the parent candidate substitutes each word in their respective cluster. We choose the most frequent word variation as the parent candidate of the cluster. Applying this shows a significant reduction in total unique tokens, and an increase in the frequency of remaining words. Due to this reduction in count of unique tokens by removing the noisy word variations, we demonstrate NLP Tasks performed on the new dataset to be more reliable.

## 6 Experiments and Evaluation

We tested our approach on the following tasks to evaluate our model

- **Sentiment Analysis on CSMT**: [12] have learned sub-word level representations in LSTM (Subword-LSTM) architecture for sentiment analysis on Hindi-English CSMT. They applied character and sub-word based networks to classify given CSMT text into three sentiments classes - positive, negative and neutral. Performing the same task but along with our preprocessing approach on the data improved the results ( illustrated in table 5 ).
- **POS-tagging on CSMT**: We tested our preprocessing approach on POS-tagging of CSMT text [4] Hindi-English language pair. [4] used Conditional Random Field, a sequence learning method, to capture patterns of sequences containing code switching to tag each word with accurate part-of-speech information. Introduction of our preprocessing approach in [4] improved the performance ( illustrated in table 6 ).

## 7 Conclusions

In this paper, we demonstrate automatic capturing and substitution of different word variations in CSMT data. Our approach exploits the property that words and their variations share similar context in a large noisy text. We also validated our approach by using the methodology on POS-tagging and sentiment analysis of CSMT text which showed improvements over their state-of-the-art counterparts. We demonstrated that informal word variations in CSMT data belong to the same cluster of semantic space and our clustering approach helps in improving the efficiency of finding the candidate words for the substitution phase.

---

[6] https://matplotlib.org/
[7] https://lvdmaaten.github.io/tsne/

This work contributes to the initial step towards building a generic model for automated preprocessing of CSMT data that is valid across large corpora and multiple language pairs. Future extensions of this work could include testing on diverse NLP tasks on CSMT data, as well as across many more language pairs, especially exploring the challenges faced by working on language pairs beyond languages of the Indian subcontinent.